\documentclass[10pt,oneside]{article}

\usepackage{times}
\usepackage{fullpage}
\usepackage{flafter}
\usepackage{graphicx}
\date{}

\usepackage[breaklinks=true,colorlinks,bookmarks=false]{hyperref}

\begin{document}

\title{Object Tracking via Dynamic Feature Selection Processes\\\large Short Description Page: The Visual Object Tracking 2016 Challenge}

\author{Giorgio Roffo\\
University of Verona\\
Strada le Grazie 15, Verona, Italy\\
{\tt\small giorgio.roffo@univr.it}
\and
Simone Melzi\\
University of Verona\\
Strada le Grazie 15, Verona, Italy\\
{\tt\small simone.melzi@univr.it}
}

\newcommand{\replace}[1]{$\langle$\textit{#1}$\rangle$}

\maketitle

\section{Full name of the tracker}
Dynamic Feature Selection Tracker

\section{Abbreviated name of the tracker}
DFST

\section{Brief description of the method}

We propose an optimized visual tracking algorithm based on the real-time selection of locally and temporally discriminative features. A novel feature selection mechanism is embedded in the Adaptive Color Names~\cite{Danelljan_2014_CVPR} (ACT) tracking system that adaptively selects the top-ranked discriminative features for tracking. The Dynamic Feature Selection Tracker (DFST) provides a significant gain in accuracy and precision allowing the use of a dynamic set of features that results in an increased system flexibility. Our ranking solution is based on the Inf-FS~\cite{Roffo_2015_ICCV}. The Inf-FS is an unsupervised method, it ranks features according with their ``redundancy" (for further details on the original method see ~\cite{RoffoBMVC2016}). For the sake of foreground/background separation, we propose a supervised variant that is able to score high features with respect to class ``relevancy", that is, how well each feature discriminates between foreground (target) and background. Therefore, we design the input adjacency matrix of the Inf-FS in a supervised manner by significantly reducing the time needed for building the graph. The ACT tracking system does not fit the size of the bounding box. Indeed, in the original framework, the bounding box remains of the same size during the tracking process. We propose a simple yet effective way of adapting the size of the box by using a fast online algorithm for learning dictionaries~\cite{Mairal:2009}. At each update, we use multiple examples around the target (at different positions and scales), we find tight bounding boxes enclosing the target by selecting the one that minimizes the reconstruction error. Thus, we also improved the ACT by adding micro-shift at the predicted position and bounding box adaptation.

\section{The Adaptive Color Tracking System}\label{sec:ACTtracker}\vspace{-0.05cm}

The ACT system~\cite{Danelljan_2014_CVPR} is a recent solutions for tracking that do not apply pre-learned models of object appearance. ACT actually extends the CSK tracker with color information obtained from color names~\cite{VSVL09} (CNs) of a target object and learn an adaptive correlation filter by mapping multi-channel features into a Gaussian kernel space. Schematically, the ACT tracker contains three improvements to CSK tracker: (1) A temporally consistent scheme for updating the tracking model is applied instead of training the classifier separately on single samples, (2) CNs are applied for image representation, and (3) ACT employs a dynamically adaptive scheme for selecting the most important combinations of colors for tracking.

In the ACT framework, for the current frame $p$, CNs are extracted and used as features for visual tracking. Moreover, a grayscale image patch is preprocessed by multiplying it with a Hann window, then, the final representation is obtained by stacking the luminance and color channels. 
The ACT algorithm considers all the extracted appearances $x^p$ to estimate the associated covariance matrix $C_p$. A projection matrix $B_p$, with orthonormal column vectors, is calculated by eigenvalue decomposition (EVD) over $C_P$. 

Let $x^p_1$ be the $D_1$-dimensional learned appearance. Thus, the projection matrix $B_p$ is used to compute the new $D_2$-dimensional feature map $x^p_2$ of the appearance by the linear mapping $x^p_1(m,n) = B^T_px^p_2(m,n), \forall m,n$. The projection matrix $B_p$ is updated by selecting the $D_2$ normalized eigenvectors of $R_p$ (see Eq.\ref{eq:ATC1}), that corresponds to the largest eigenvalues. 
\begin{equation}\label{eq:ATC1}
	R_p = C_p + \sum^{p-1}_{j=1}B_j \Lambda_jB^T_j,
\end{equation}
where $C_p$ is the covariance matrix of the current appearance and $\Lambda_j$ is a $D_2 \times D_2$ diagonal matrix of weights needed for each basis in $B_p$. Finally, $D_2$ represents the number of dimensions where the $D_1$ features are projected on.

Summarizing, $B_p$ is used to transform the original feature space (i.e., a fixed set of features which comprises the 10 CNs) to yield a subspace by performing dimensionality reduction (i.e., the first 2 principal components). Finally, the projected features plus the gray-scale patch of the target are used to compute the detection scores and the target position in in the new frame $p+1$ by  cross correlation (see ~\cite{RoffoBMVC2016,Danelljan_2014_CVPR} for further details).

\section{The Proposed Method}

This section describes the proposed solution which can be mainly divided into two parts. Firstly, the tracker ranks the set of features dynamically at each frame. Secondly, it selects a subset and updates the current set of features used for tracking.

\subsection{Step 1: On-Line Feature Selection}

The first step of our approach is feature selection. The ACT system uses a set of $10$ color names for each pixel of the target box, as described in the previous section. Our ranking solution is based on the Inf-FS~\cite{Roffo_2015_ICCV}. Each feature is mapped on an affinity graph, where nodes represent features, and weighted edges the relationships between them. In the original version, the graph is weighted according to a function which takes into account both correlations and standard deviations between feature distributions in an unsupervised manner. The computation of all the weights results to be the bottleneck of this method, at least when the speed is a primary requirement. The inf-FS method is unsupervised, we propose a supervised solution which is able to score high features with respect to how well each feature discriminate between foreground (target) and background. 

Therefore, we propose a supervised version of the Inf-FS algorithm, where we significantly reduced the time needed for building the graph.

Firstly, we labeled all the pixels into the bounding box of the target as positive (+1), then we select the immediate background of the target box and we label each pixel as belonging to the negative class (-1).

Given the set of $10$-dimensional samples for classes $Target = C_1$ and $Background = C_2$, and the set of features $F = \{  f^{(1)}, ..., f^{(10)} \}$, where $f^{(i)}$ is the distribution of the $i$-th feature over the samples, we measure the tendency of the $i$-th feature in separating the two classes by considering three different measurements. 

The Fisher criterion:
 \[
	 p_i = \frac{\left | \mu_{i,1} - \mu_{i,2} \right |^2}{ \sigma_{i,1}^2+\sigma_{i,2}^2}
\]
where $\mu_{i,k}$ and $\sigma_{i,k}$ are the mean and standard deviation, respectively, assumed by the $i$-th feature when considering the samples of the $k$-th class. The higher $p_i$, the more discriminative is the $i$-th feature. 

As for a second measure for class separation, we perform a t-test of the null hypothesis that foreground samples and background samples are independent random samples from normal distributions with equal means and equal but unknown variances, against the alternative that the means are not equal. We consider a rejection of the null hypothesis at the 5\% significance level. The t-test is the following:
 \[
	 tt_i = \frac{ \mu_{i,1} - \mu_{i,2}  }{ \sqrt{ \frac{\sigma_{i,1}^2}{n_1}  +  \frac{\sigma_{i,2}^2}{n_2} }}
\]
where $n_k$ is the sample size, and $tt_i$ is the p-value, that is, the probability, under the null hypothesis, of observing a value as extreme or more extreme of the test statistic. In our case, the test statistic, under the null hypothesis, has Student's t distribution with $n_1 + n_2 - 2$ degrees of freedom. 

Finally, we use the Pearson's linear correlation criterion $c_i$. The central hypothesis is that good feature sets contain features that are highly correlated with the class.

Since we are interested in identifying \emph{subsets} of features which are maximally discriminative, we continue our analysis by considering pairs of features. To this sake, we firstly create a final vector $s$ by averaging all the previous feature evaluation metrics, then we compute the pairwise matrix $A = s \cdot s^\top$. The generic entry $a_{ij}$ accounts for how much discriminative are the feature $i$ and $j$ when they are jointly considered; at the same time, $a_{ij}$ can be considered as a weight of the edge connecting the nodes $i$ and $j$ of a graph, where the $i$-th node models the $i$-th feature distribution. Under this perspective, the matrix $A$ models a fully connected graph, where paths of arbitrary length can encode the discriminative power of a set of features by simply multiplying the weights of the edges that form them.\\

The matrix $A$ is fed to the inf-FS approach. Therefore, the Inf-FS evaluates each path on the graph defined by $A$ as a possible selection of features. Letting these paths tend to an infinite length permits the investigation of the relevance of each possible subset of features to class labels. The supervised Inf-FS assigns a final score to each feature of the initial set; where the score is related to how much a given feature is a good candidate for foreground/background separation. Therefore, ranking the outcome of the Inf-FS in descendant order allows us to reduce the number of features and to select only the most relevant for the current frame.

\subsection{Step 2: Embedding Feature Selection}

\begin{figure}
\center
\includegraphics[width=0.98\textwidth]{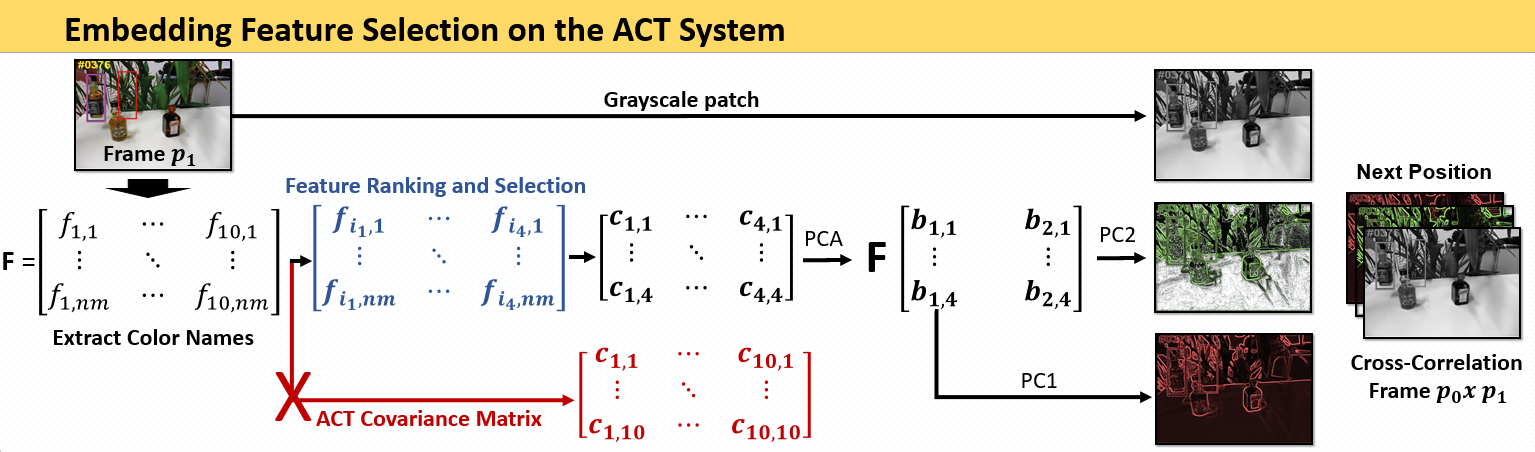}
\caption{Pipeline: Embedding Feature Selection on the ACT System }
\label{fig:frames}
\end{figure}

The supervised Inf-FS algorithm is suitable to be embedded on the ACT system Fig.~\ref{fig:frames}. Firstly, the ACT appearances are extracted for object $x_{pos}$ and background $x_{neg}$ classes, and computed using samples taken from the most recently tracked frame. Secondly, feature ranking is applied. This important step can be interpreted as ranking by relevance the dimensions of the feature vector $x_{pos}$, where features in the first ranked positions are the ones that better discriminate the object from the background in the current frame. Finally, these features are used to estimate the  covariance matrix $C$ of the ACT, resulting in an impressive improvement of the baseline tracker.

%
 
\section{Comparison with the ACT system}\label{sec:Comparison}

Tracking success or failure depends on how a tracking system can efficiently distinguish between the target and its surroundings. ACT use a fixed set of features that results in a reduced system flexibility. Intrinsic and extrinsic factors affect the target appearance by introducing a strong variability. Our solution allows to deal with drastic changes in appearance and to maintain good execution time (i.e., high frame rates). The ACT tracker does not take into account the appearance of the background, thus, does not appropriately manage the presence of strong distractors. ACT enhanced by feature selection penalizes features that produces spatially-correlated background clutter or distractors. The Inf-FS also allows to deal with strong illumination variation, by selecting those features which remain different when affected by the light.

\section{Features used}

In the DFST framework, for the current frame $p$, color names~\cite{VSVL09} are extracted and used as features for visual tracking. Moreover, a grayscale image patch of the target is preprocessed by multiplying it with a Hann window, then, the final representation is obtained by stacking the luminance (target patch) and color channels.

\section{The VOT2016 challenge results}
This tracker was evaluated according to the rules of the VOT2016 challenge
as specified in the VOT2016 evaluation kit document~\cite{VOT2016}~\cite{kristan2016visual}.
The authors guarantee that they have exactly followed the guidelines
and have not modified the obtained results in any way that would violate the challenge rules.

\subsection{Experimental environment}

During the experiments, the tracker was using neither GPU nor any kind of distributed processing.\newline

  \begin{itemize}
    \item The platform used is a PCWIN64 (Microsoft Windows 10), on an Intel i7-4770 CPU 3.4GHz 64-bit, 16.0 GB of RAM, using MATLAB ver.2015a.
    \item The processing speed measured by the evaluation kit is up to $16.53$ (baseline), $5.06$ (unsupervised).
    \item  As for the main parameters we set for the challenge, we highlight the learning rate for appearance model update scheme up to 0.005 which gives more importance to the appearance learnt in the first frames. The extra area surrounding the target box (padding) is automatically estimated according with the ration between the frame size and target bounding box size. We also reduce the learning rate for the adaptive dimensionality reduction to 0.1. The number of selected features is up to 8, and the dimensionality of the compressed features is 4. As for the dictionary learning, DFST learns a dictionary with 250 elements, by using at most 200 iterations. 
  \end{itemize}

    {\small
      \bibliographystyle{abbrv}
      \bibliography{results}
    }

    \end{document}